\pdfoutput=1

\documentclass[11pt]{article}

\usepackage[final]{acl}

\usepackage{times}
\usepackage{latexsym}
\usepackage{bm}
\usepackage[linesnumbered,vlined,ruled,commentsnumbered]{algorithm2e}

\usepackage[T1]{fontenc}

\usepackage[utf8]{inputenc}

\usepackage{microtype}

\usepackage{inconsolata}

\usepackage{graphicx}
\usepackage{amssymb}

\usepackage{amsthm,amsmath}
\usepackage{mathrsfs}

\usepackage{amsmath}
\usepackage{multirow}
\usepackage{enumerate}
\usepackage{enumitem}
\usepackage{booktabs}
\usepackage{array}
\usepackage{makecell}
\usepackage{colortbl}
\usepackage{hhline}

\usepackage{subcaption}
\usepackage{tabularx}
\usepackage{rotating}

\title{Emotion Transfer with Enhanced Prototype for Unseen Emotion Recognition in Conversation}

\author{
 \textbf{Kun Peng\textsuperscript{1,2}},
 \textbf{Cong Cao\textsuperscript{1}},
 \textbf{Hao Peng\textsuperscript{3}},
 \textbf{Guanlin Wu\textsuperscript{4}},
 \textbf{Zhifeng Hao\textsuperscript{5}},
 \textbf{Lei Jiang\textsuperscript{1}},\\
 \textbf{Yanbing Liu\textsuperscript{1,2}},
 \textbf{Philip S. Yu\textsuperscript{6}},
\\
\\
 \textsuperscript{1}Institute of Information Engineering, Chinese Academy of Sciences, China
 \\
 \textsuperscript{2}School of Cyber Security, University of Chinese Academy of Sciences, China
 \\
 \textsuperscript{3}Beihang University, China
 \textsuperscript{4}National University of Defense Technology, China\\
 \textsuperscript{5}Shantou University, China
 \textsuperscript{6}University of Illinois at Chicago, USA\\
 \small{
   \textbf{Correspondence:} \href{pengkun@iie.ac.cn}{\{pengkun, caocong\}@iie.ac.cn}
}
}

\begin{document}
\maketitle
\begin{abstract}
Current Emotion Recognition in Conversation (ERC) research follows a closed-domain assumption. However, there is no clear consensus on emotion classification in psychology, which presents a challenge for models when it comes to recognizing previously unseen emotions in real-world applications. To bridge this gap, we introduce the Unseen Emotion Recognition in Conversation (UERC) task for the first time and propose \textbf{ProEmoTrans}, a solid prototype-based emotion transfer framework. This prototype-based approach shows promise but still faces key challenges: 
First, implicit expressions complicate emotion definition, which we address by proposing an LLM-enhanced description approach. Second, utterance encoding in long conversations is difficult, which we tackle with a proposed parameter-free mechanism for efficient encoding and overfitting prevention. Finally, the Markovian flow nature of emotions is hard to transfer, which we address with an improved Attention Viterbi Decoding (AVD) method to transfer seen emotion transitions to unseen emotions.
Extensive experiments on three datasets show that our method serves as a strong baseline for preliminary exploration in this new area.
\end{abstract}

\section{Introduction}
\begin{figure}[t]
    \centering
    \begin{minipage}{0.25\textwidth}
        \centering
        \includegraphics[width=\textwidth]{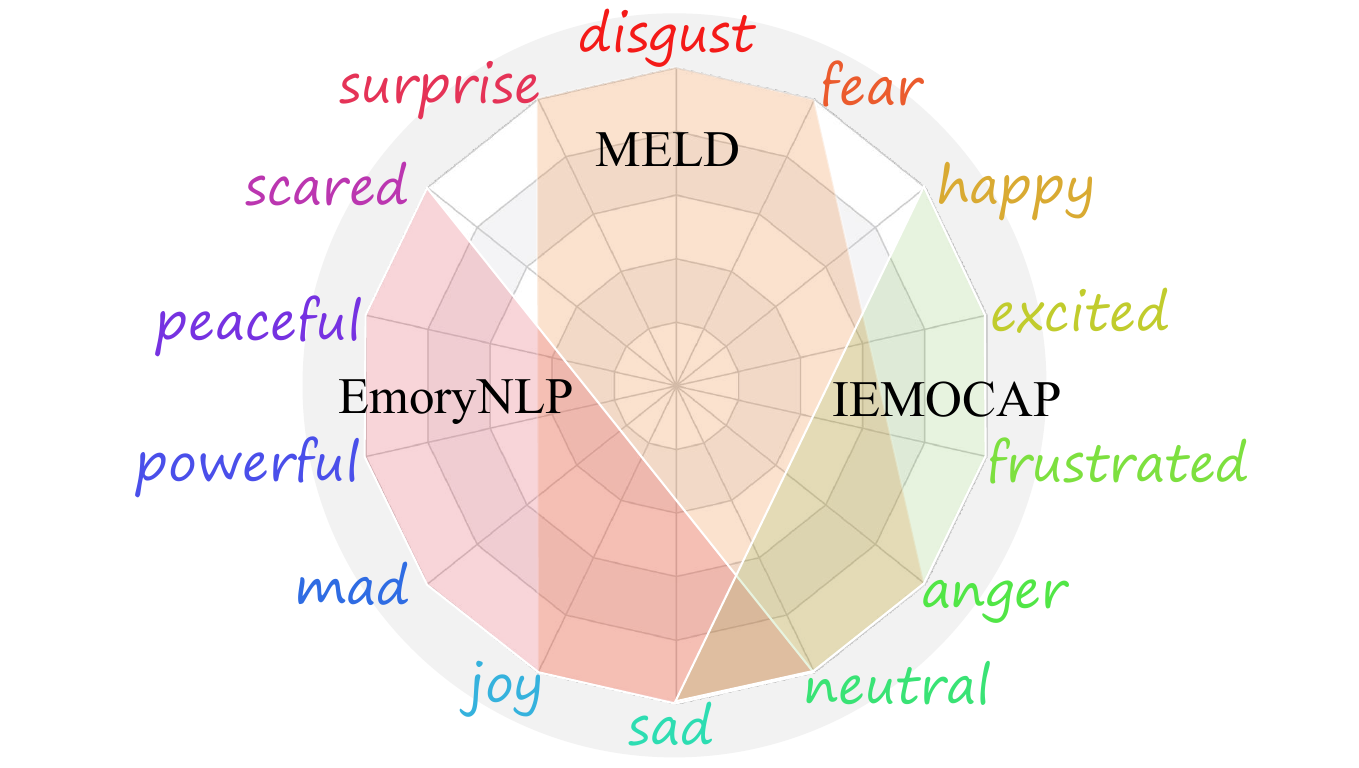}
        \vspace{-0.5cm}
        \subcaption{The emotion categories.}
    \end{minipage}%
    \begin{minipage}{0.25\textwidth}
        \centering
        \includegraphics[width=\textwidth]{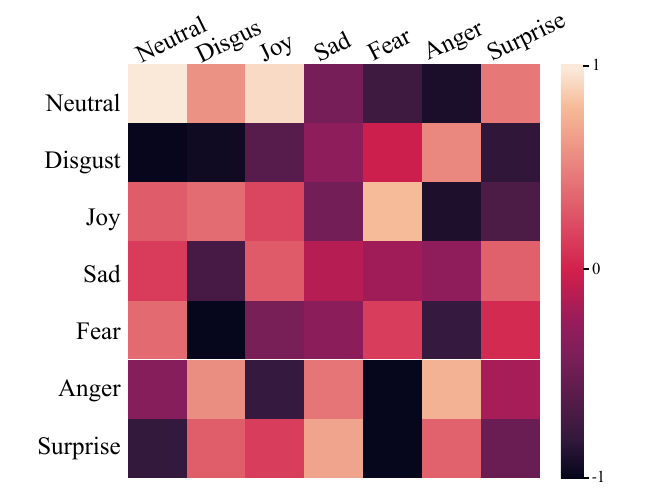}
        \vspace{-0.5cm}
        \subcaption{Transition scores.}
    \end{minipage}
    \caption{(a) shows that the emotion categories in three foundational datasets vary significantly in the emotion labels. (b) shows the transition scores learned on the MELD dataset.}
    \label{fig:0}
\end{figure}

Emotion Recognition in Conversation (ERC) aims to predict the emotional state of each utterance in multi-turn conversations, holding significant research value in areas such as Conversational Sentiment Analysis \cite{li-etal-2023-diaasq} and Empathetic Responses \cite{ijcai2022p600}.
However, in the field of psychology, existing research works \cite{ekman19991, plutchik2013theories, cowen2017self} feature a variety of emotion classification theories, yet they have not reached a clear consensus\footnote{For instance, \citet{plutchik2013theories} categorizes emotions into 32 types, while \citet{cowen2017self} categorizes emotions into 27 types, including unusual emotions like \textit{nostalgia} and \textit{sexual desire}.}. 
Due to the complex definitions and the various classification theories, in real-world applications, such as open-domain dialogue systems, it is likely to occur new emotions that are unseen in the training stage.
As shown in Figure \ref{fig:0} (a), the emotion labels across three widely used datasets \cite{busso2008iemocap,zahiri2018emotion,poria-etal-2019-meld} exhibit significant non-overlapping portions. 
This makes it challenging to directly apply models trained on a single dataset to other datasets. 
For instance, a model trained on the MELD dataset may struggle to recognize the emotion \textit{powerful} in the EmoryNLP dataset.

To bridge this gap, we introduce the Unseen Emotion Recognition in Conversation (UERC) task for the first time, which aims to predict unseen emotions by leveraging prior knowledge from seen emotions in training data.
To address this task, we attempt the prototype-based approaches \cite{chen-li-2021-zs,zhao-etal-2023-matching,li-etal-2024-alignre} to learn a prototype vector for each emotion, helping the model capture the distinct meaning of emotions. However, three key challenges hinder progress.
\textbf{Challenge 1: Implicit emotion expression}. 
Existing methods primarily rely on the provided label descriptions to enhance prototype semantics. 
However, the UERC task lacks emotion descriptions, and, even more critically, many complex emotions are hard to define clearly, and relying solely on descriptive information is insufficient to obtain robust and faithful prototypes.
\textbf{Challenge 2: Hard utterance encoding}. Due to the extensive length of conversation texts, 
existing ERC methods \cite{majumder2019dialoguernn,hu-etal-2021-dialoguecrn,zhang-etal-2023-dualgats,10418539} typically follow two steps: encoding utterance representations first, then modeling inter-utterance features with additional relation-learning modules.
However, our preliminary experiments indicate that these additional modules can lead to overfitting the training data, compromising the model's ability to generalize to unseen emotions.
Conversely, removing these modules results in losing valuable inter-utterance relations, creating a dilemma.
\textbf{Challenge 3: Unadapted emotion transition}. It's found that emotions exhibit a Markov property \cite{DBLP:conf/icassp/SongZZHH22}, whereby the current utterance's emotion is influenced by preceding ones.
As illustrated in Figure \ref{fig:0} (b), when the current emotion is \textit{Disgust}, the transfer score for \textit{Anger} in the subsequent utterance is notably highest, aligning with intuitive expectations. 
While the Markov property can effectively aid emotion prediction, the transfer score matrix for unseen emotions cannot be pre-learned.

To address these challenges, we propose a solid prototype-based emotion transfer framework called \textbf{ProEmoTrans}. 
Specifically, to address the \textbf{implicit emotion expression} challenge, we first employ a dictionary to obtain all the emotion descriptions. 
We then leverage the in-context learning capabilities of large language models (LLMs) to generate utterances that implicitly express these emotions, thereby enhancing the model's comprehension of complex emotions.
To address the \textbf{hard utterance encoding} challenge, we refrain from using additional relation-learning modules to prevent the model from overfitting to seen emotions. 
Instead, we propose a Gaussian Self-Attention mechanism to capture inter-utterance relations. 
This parameter-free mechanism obtains utterance embeddings by using linear combinations of contextual representations, effectively leveraging relation information among utterances at varying distances.
To leverage the \textbf{emotion transition}, we propose an improved Attention Viterbi Decoding (AVD) algorithm within the Conditional Random Field (CRF) framework, enabling the capture of transition probabilities for seen emotions between all adjacent utterances. 
Subsequently, we extend the transition probabilities of seen emotions to unseen emotions by utilizing prototype similarity.
Our contributions can be summarized as follows:


1) We propose the UERC task for the first time and introduce a novel model called ProEmoTrans\footnote{Available at https://github.com/KunPunCN/ProEmoTrans/}. Extensive experiments on three datasets demonstrate that this method serves as a solid baseline.

2) We leverage the prior knowledge of LLMs to generate implicit contexts that enhance complex emotion prototypes. 

3) We introduce a Gaussian self-attention mechanism that effectively utilizes inter-utterance relations while avoiding overfitting to seen emotions. 

4) We improve the Viterbi decoding algorithm to extend the transition probabilities of seen emotions to unseen emotions.

\section{Related Work}

\subsection{Emotion Recognition in Conversation}
ERC in a text-modality setting is an active research topic. Early RNN-based \cite{jiao-etal-2019-higru,majumder2019dialoguernn,hu-etal-2021-dialoguecrn} and GCN-based \cite{ghosal-etal-2019-dialoguegcn,shen-etal-2021-directed,zhang-etal-2023-dualgats} methods tried to model the temporal features or conversational structures.
Some other studies \cite{ghosal-etal-2020-cosmic,ong2022discourse} have also attempted to integrate more common-sense knowledge.
The latest contrastive-based methods \cite{hu-etal-2023-supervised,10040720,yu-etal-2024-emotion} focus on using contrastive learning to distinguish semantically similar emotions.
While these additional modules can effectively help the model fit the distributions of seen emotions, in the UERC setting, they can impair the model's ability to generalize to unseen emotions.

\begin{figure*}[t]
    \centering
    \includegraphics[width=1\textwidth]{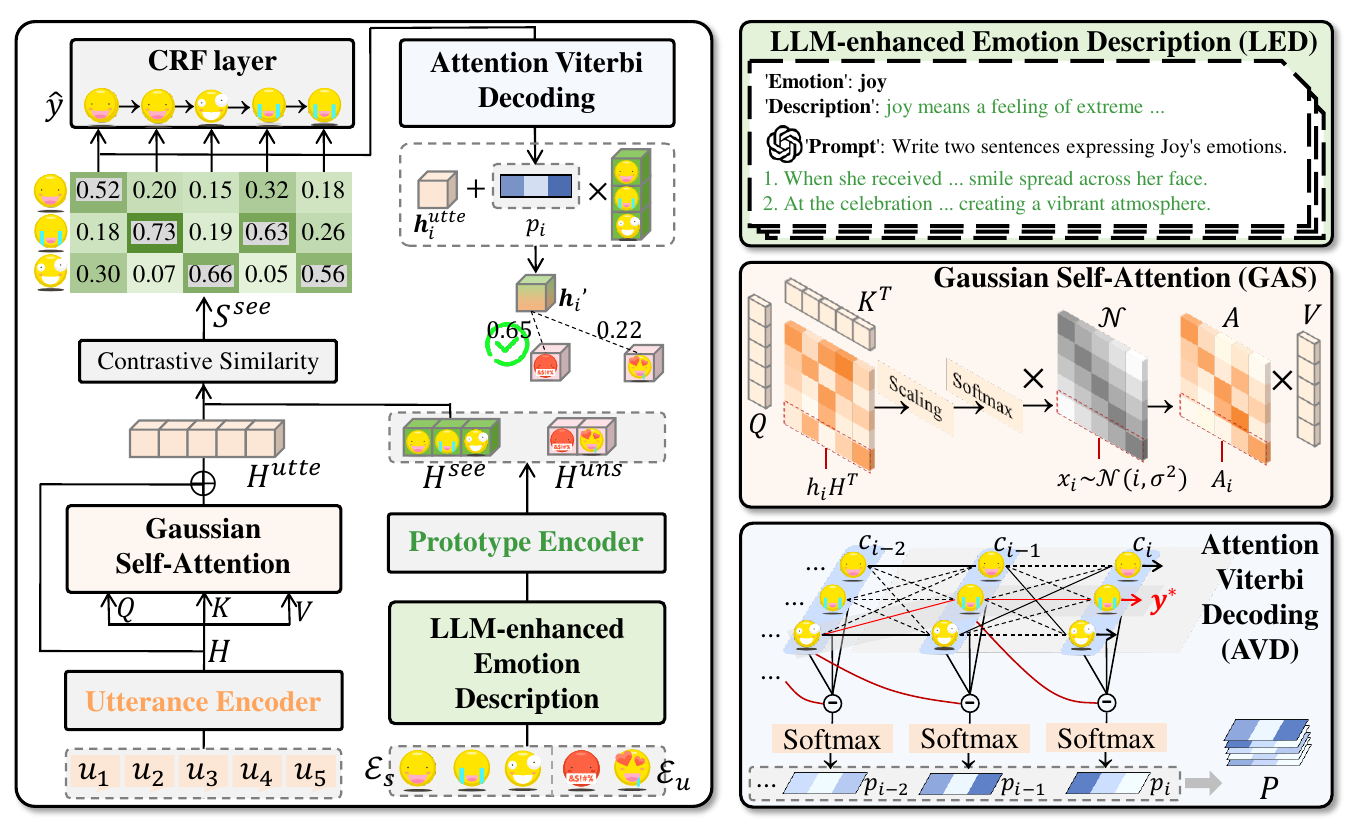}
    \vspace{-0.5cm}
    \caption{The architecture of our proposed EmoTrans.}
    \label{fig:2}
\end{figure*}

\subsection{Zero-shot Learning in ERC}
Zero-shot Learning (ZSL) aims to train a model on one label set and then apply it to another set of previously unseen labels. 
Currently, research on ZSL in the ERC field is quite limited. 
A work that is closely related to ours is CTPT \cite{xu-etal-2023-efficient}, which focuses on cross-task few-shot settings, while we are the first to explore the model's ability in zero-shot predicting for unseen emotions.
In the zero-shot setting, CTPT primarily improves the recognition of similar emotions across tasks but performs poorly in recognizing unseen emotions.

Prototype alignment \cite{chen-li-2021-zs,zhao-etal-2023-matching,li-etal-2024-alignre} is a powerful method in ZSL. It first encodes sentence and label information into a hidden vector space, then aligns sentence embeddings with label prototype embeddings using semantic matching. 
Through this process, the model acquires the ability to generalize label knowledge. 
During the inference phase, the model encodes the unseen label information and makes predictions through nearest neighbor search.
In other prototype-based zero-shot NLP research fields, such as zero-shot relation extraction, \citet{zhao-etal-2023-matching} proposes a fine-grained semantic matching method to reduce the negative impact of irrelevant features. \citet{li-etal-2024-alignre} enhances label prototypes by introducing more side descriptions.
\section{Methodology}
\subsection{Task Definition}
Given a conversation $\mathcal{U}=\{(u_1,t_1), (u_2,t_2), ...,$ $(u_N,t_N)\}$, where each utterance $u_i$ has only one speaker $t_i$, $N$ is the total number of utterances. Different utterances may belong to the same speaker, so it is possible to have $t_i = t_j (i \neq j)$.
%
The training dataset $\mathcal{D}_s$ has a set of seen emotions $\mathcal{E}_s$, and the test dataset $\mathcal{D}_u$ has a set of unseen emotions $\mathcal{E}_u$.
%
There is no overlap between $\mathcal{E}_s$ and $\mathcal{E}_u$.
The objective of the UERC task is to learn from $\mathcal{D}_s$ and transfer the model to predict the unseen emotion label $e^{uns}_i \in \mathcal{E}_u$ of each utterance $u_i$.


\subsection{Framework of ProEmoTrans}
The overall architecture of our proposed ProEmoTrans is illustrated in Figure \ref{fig:2}.

\subsubsection{Emotion Prototype Encoding}

Given a seen emotion word $e^{see}_i \in \mathcal{E}_s$, we can find its corresponding description\footnote{All the descriptions are listed in Appendix~\ref{sec:appendix}.} $X^{desc}_i$ in the Wiktionary\footnote{https://en.m.wiktionary.org}.
However, unlike direct descriptions, emotions in conversation are often expressed implicitly.
This gap makes the prototype learned from emotional descriptions lack sufficient generalization, especially for more complex emotions (e.g., powerful).

To improve the quality of emotion prototypes, we propose the \textbf{LLM-enhanced Emotion Description (LED)} method.
We first design a prompt template \textit{Write two sentences expressing} [\textit{MASK}]\textit{'s emotions.} 
Afterward, by filling in the [\textit{MASK}] position with the emotion word $e^{see}_i$ and leveraging the LLM's prompt generation capabilities, we generate sentences $X^{llm}_i$ that implicitly express that emotion. 
%
The enhanced description $X^{see}_i$ is defined as the concatenation of $e^{see}_i$, $X^{desc}_i$ and $X^{llm}_i$:
\begin{equation}
    X^{see}_i = \{[\textit{CLS}], e^{see}_i, X^{desc}_i, X^{llm}_i, [\textit{SEP}]\}.
\end{equation}
We feed it into the prototype encoder to obtain the final emotion prototype $\bm{h}^{see}_i$:
\begin{equation}
    \bm{h}^{see}_i = Encoder_E(X^{see}_i)[0],
\end{equation}
where $\bm{h}^{see}_i \in \mathbb{R}^d$ is the first token (i.e., [\textit{CLS}]) of the last hidden layer.
Through the above process, we can encode the prototypes of each emotion word in the seen emotion set $\mathcal{E}_s$ and obtain $\bm{H}^{see} = (\bm{h}^{see}_1,\bm{h}^{see}_2,...,\bm{h}^{see}_n )$.
Similarly, for the unseen emotion set $\mathcal{E}_u$, we have $\bm{H}^{uns} = (\bm{h}^{uns}_1,\bm{h}^{uns}_2,...,\bm{h}^{uns}_m)$, where $n$ and $m$ are the numbers of emotions in $\mathcal{E}_s$ and $\mathcal{E}_u$, respectively.

\subsubsection{Utterance Encoding}
\label{section322}
Following previous works \cite{hu-etal-2021-dialoguecrn,shen-etal-2021-directed,zhang-etal-2023-dualgats}, due to the conversation text being too lengthy, we use an utterance encoder to obtain the utterance representation $\bm{h}_i$:
\begin{equation}
    \bm{h}_i = Encoder_U(u_i)[0].
    \label{eq3}
\end{equation}
The representation of all utterances is denoted as $\bm{H} \in \mathbb{R}^{N \times d}$, where $N$ is the number of utterances in $\mathcal{U}$.
After that, we propose a non-parametric \textbf{Gaussian Self-Attention (GSA)} mechanism that effectively learns the inter-utterance relationships and alleviates overfitting to seen emotions.

Given the token $\bm{h}_i \in \bm{H}$, the Gaussian attention score $\bm{A}_i \in \mathbb{R}^N$ that attends to $\bm{H}$ is defined as:
\begin{equation}
    \bm{A}_i = Softmax(\frac{\bm{h}_i\bm{H}^T}{d})\bm{\mathcal{N}}_i,
\end{equation}
where $\bm{\mathcal{N}}_i \in \mathbb{R}^N$ are discrete values that follow the Gaussian distribution $\mathcal{N}(i, \sigma^2)$, and the variance $\sigma$ is a hyperparameter.
Using the Gaussian attention score, we aggregate highly relevant information from the entire conversation while reducing the impact of distant tokens. 
This inter-utterance relationship aggregation follows a non-parametric linear operation:
\begin{equation}
    \bm{h}^{utte}_i = \bm{h}_i + \bm{A}_{i}\bm{H},
\end{equation}
where $\bm{h}^{utte}_i \in \mathbb{R}^{d}$ is the updated utterance representation.
The final representation of all utterances is denoted as $\bm{H}^{utte} \in \mathbb{R}^{N \times d}$.

The GSA mechanism has two key properties: First, parameter-free. Previous supervised methods used parameterized modules (such as LSTM and GCN) to learn inter-utterance relationships. However, in unsupervised scenarios, parameterized modules led to overfitting on the training set, hindering generalization on unseen datasets (Appendix \ref{sec:appendixa1}). 
Second, distance-aware learning of inter-utterance relationships. Directly sampling discrete values from a one-dimensional Gaussian distribution based on the distance between utterances, with closer utterances receiving more attention.

\subsubsection{Contrastive Similarity and Training}
In the above sections, we obtained emotion prototypes and utterance representations. 
In this section, through nearest neighbor search, we can align utterances with their corresponding emotion labels.
Inspired by infoNCE \cite{oord2018representation}, we define a contrastive similarity to pull the utterance embeddings closer to their corresponding prototype embeddings while pushing apart the inconsistent ones.
This similarity $\bm{S}^{see} \in \mathbb{R}^{N \times n}$ is defined as:
\begin{equation}
    \bm{S}^{see} = Sim(\bm{H}^{utte},\bm{H}^{see}),
    \label{eq6}
\end{equation}
\begin{equation}
    s^{see}_{ij} = \frac{e^{cos(\bm{h}^{utte}_i,\bm{h}^{see}_j)/\tau}}{\sum\nolimits_{j=1}^{n}e^{cos(\bm{h}^{utte}_i,\bm{h}^{see}_j)/\tau}},
    \label{eq7}
\end{equation}
where Eq. (\ref{eq7}) is the details of Eq. (\ref{eq6}).
$cos(\cdot)$ is a cosine similarity function and $\tau$ is a temperature hyperparameter. 
$s^{see}_{ij}$ represents the probability of the $i$-th utterance expressing the $j$-th seen emotion.

Due to the transition dependencies between emotions, independent predictions are insufficient. Therefore, we subsequently feed $\bm{S}^{see}$ into a Conditional Random Field (CRF) \cite{lafferty2001conditional}.
For a sequence of predictions: $\bm{y} = (y_1, y_2, ..., y_N)$, its CRF score can be defined as:
\begin{equation}
    \mathcal{C}(\bm{y}) = \sum\nolimits_{k=0}^{N}\bm{M}_{y_k,y_{k+1}} + \sum\nolimits_{k=1}^{N}\bm{S}^{see}_{k,y_k},
    \label{eq8}
\end{equation}
where $\bm{M} \in \mathbb{R}^{(n+2) \times n}$ is the transition matrix\footnote{$n+2$ is because there are start and end transitions here.} of the CRF layer. $y_0$ and $y_{N+1}$ are the additional \textit{start} and \textit{end} tags.
The probability of the sequence $\bm{y}$ is a softmax over the scores of all possible sequences:
\begin{equation}
 p(\bm{y}) = \frac{e^{\mathcal{C}(\bm{y})}}{\sum\nolimits_{\tilde{y} \in \bm{Y_\mathcal{U}}}e^{\mathcal{C}(\bm{\tilde{y}})}},
\end{equation}
where $\bm{Y_\mathcal{U}}$ represents all possible predicted sequences.
Our training goal is to minimize the loss: $\mathcal{L} = - log(p(\bm{\widehat{y}}))$, where $\bm{\widehat{y}}$ represents the true sequences.

\subsubsection{Inference}
The original Viterbi decoding is limited to the seen emotions, and the valuable emotion transition dependencies learned by the CRF layer cannot be adapted to unseen emotions.
To address this gap, we propose the \textbf{Attention Viterbi Decoding (AVD)} algorithm.
We define the score of the $i$-th utterance expressing the $j$-th seen emotion as:
\begin{equation}
    c_{ij} = \mathop{\max}\limits_{\bm{\tilde{y}} \in \bm{Y}_{\bm{\mathcal{U}}_{[1:i]}}, \tilde{y}_k=j}{\mathcal{C}(\bm{\tilde{y}})},
    \label{eq11}
\end{equation}
where $c_{0j}=\bm{M}_{0,j}$. 
$\bm{Y}_{\bm{\mathcal{U}}_{[1:i]}}$ represents all possible tag sequences from $u_1$ to $u_i$.
The score $c_{ij}$ represents the maximum CRF score of all possible sequences ending with $\tilde{y}_i = e^{see}_j$.
Based on Eq. (\ref{eq8}), we can derive that:
\begin{equation}
    c_{ij} = \mathop{\max}\limits_{1 <= k <= n}{(c_{(i-1)k}+\bm{M}_{k,j}+\bm{S}^{see}_{k,j})}.
    \label{eq12}
\end{equation}
The time complexity of calculating a single $c_{ij}$ is $\mathcal{O}(n)$. The overall time complexity for traversing all $c_{ij}$ is $\mathcal{O}(Nn^2)$.
During the traversal, we also record the path $\bm{y}^*=(y^*_1,...,y^*_N)$ with the maximum CRF score, such that $c_{Ny^*_{N}} > c_{Nj}, \forall j \neq y^*_{N}$.

The final output of the AVD algorithm is a probability matrix $\bm{\mathcal{P}} \in \mathbb{R}^{N \times n}$, where each $p_{ij} \in \bm{\mathcal{P}}$ is defined as follows:
\begin{equation}
    p_{ij} = \frac{c_{ij}-c_{(i-1)y^*_{i-1}}}{\sum\nolimits_{k=1}^{n}(c_{ik}-c_{(i-1)y^*_{i-1}})},
    \label{eq14}
\end{equation}

where $p_{ij}$ denotes the probability of the $k$-th utterance expressing the $j$-th seen emotion.
Then, we can enhance the original utterance representation using the seen emotion prototypes:
\begin{equation}
    \bm{h}'_i = \bm{h}^{utte}_i + \sum\nolimits_{j=1}^{n}p_{ij}\bm{h}^{see}_j,
\end{equation}
where $\bm{h}'_i$ incorporates the seen emotion prototypes after considering similarity (from $\bm{S}^{see}$) and emotional dependencies (from $\bm{M}$).

For a given $u_i$, the predicted unseen emotion label is obtained through nearest neighbor search:
\begin{equation}
    y^{uns}_i = \mathop{\arg\max}\limits_{1 <= j <= m}cos(\bm{h}'_i,\bm{h}^{uns}_j).
    \label{eq15}
\end{equation}

\section{Experiments Settings}

\begin{table}[h]
\centering \small
\tabcolsep=1.5pt
\begin{tabular}{c|ccc|ccc|c}
\Xhline{0.9pt}
\multirow{2}{*}{\textbf{Dataset}} & \multicolumn{3}{c|}{\textbf{\# Conversations}} & \multicolumn{3}{c|}{\textbf{\# Uterrances}}  & \multirow{2}{*}{\textbf{\# Emos.}}\\
                                  & train      & dev         & test             & train      & dev      & test            &         \\ \hline
IEMOCAP ($\bm{\mathcal{I}}$)                           & 100    & 20          & 31               & 4810     &1000        & 1623        & 6\\
EmoryNLP ($\bm{\mathcal{E}}$)                         & 659     &89           & 79               & 7551     &954        & 984         & 7                       \\
MELD ($\bm{\mathcal{M}}$)      & 1038      &114         & 280              & 9989      &1109       & 2610    &7                      \\ \Xhline{0.9pt} 
\end{tabular}
\caption{Statistics of experimental datasets.}
\label{tab:statistics}
\end{table}

\subsection{Datasets}
We evaluate our ProEmoTrans on three widely used datasets: 
\textbf{IEMOCAP} \cite{busso2008iemocap} is based on two actors performing a script. \textbf{EmoryNLP} \cite{zahiri2018emotion} and \textbf{MELD} \cite{poria-etal-2019-meld} contain scripts collected from the \textit{Friends} TV series. 
We only use the text modality of these datasets and follow previous work in splitting the IEMOCAP dataset into training and validation sets.
The dataset statistics are drawn in Table \ref{tab:statistics}.
We denote these datasets as $\bm{\mathcal{I}}$, $\bm{\mathcal{E}}$, and $\bm{\mathcal{M}}$, respectively. 
We iterate through different source datasets to train the model and use the validation and test sets of the other two datasets as the target unseen emotion datasets.
For instance, to evaluate the model trained on $\bm{\mathcal{I}}$ for its performance on $\bm{\mathcal{M}}$ test set, we select $\bm{\mathcal{E}}$ as the validation set.
The statistics of the unseen emotions under different source and target settings are shown in Appendix \ref{sec:app-dataset}.

\subsection{Implementation Details}
We utilize Bert-base-uncased \cite{DBLP:conf/nips/VaswaniSPUJGKP17} as both the utterance and prototype encoder. 
We use ChatGPT-3.5 to generate enhanced emotion descriptions.
In each training batch, we input the emotion descriptions and the utterances into the encoders simultaneously.
We use the AdamW optimizer \cite{DBLP:journals/corr/KingmaB14} with a batch size of 4 and a learning rate of $2e-5$.
The model is trained for 10 epochs with 100 warm-up steps.
All experiments are conducted with an NVIDIA RTX 8000.
The variance $\sigma$ of the Gaussian distribution is set to 0.5, and the temperature $\tau$ in Eq. (\ref{eq7}) is set to 0.02.
We use the weighted-averaged F1 score as the evaluation metric, considering only unseen emotions.
In each epoch, we evaluate the training model on the validation set and save the best one to test.
All results are averaged across five runs with different random seeds.

\subsection{Baselines}
\label{sec:baseline}
Due to limited research, we choose the following four types of baselines and make necessary modifications to their original architectures to achieve zero-shot prediction capability:

Feature-based models: \textbf{DialogueGCN} \cite{ghosal-etal-2019-dialoguegcn}, \textbf{DialogueCRN} \cite{hu-etal-2021-dialoguecrn}, and \textbf{DualGAT} \cite{zhang-etal-2023-dualgats} design special GNN/RNN-based modules to extract better utterance features and use a label-wise classification head to predict the label of each utterance. They use cross-entropy loss computed from the prediction logits and the labels.
To enable zero-shot prediction capability, we replace the classification head with a prototype encoder, which enables the model to learn prototype vectors.
Then we substitute the original cross-entropy loss with a contrastive loss based on prototype similarity (similar to Eq. \ref{eq7}).

Contrastive-based models: \textbf{SACL-LSTM} 
 \cite{hu-etal-2023-supervised}, \textbf{SCCL} \cite{10040720}, and \textbf{EACL} \cite{yu-etal-2024-emotion} focus on distinguishing semantically similar emotions using contrastive learning.
Since these models natively use representation similarity for prediction, no modifications are needed.
 
Few-shot model: \textbf{CPTC} \cite{xu-etal-2023-efficient} leverages sharable cross-task knowledge from the source task to improve few-shot performance. By removing task-specific prompts, it can also perform zero-shot prediction. Unlike in their original work, we evaluate the model only on unseen emotions. To ensure fairness, all of these comparison models use BERT-base-uncased as their backbone.

LLMs: \textbf{Llama-3.1-8b} \cite{grattafiori2024llama}, \textbf{Qwen-2.5-7b} \cite{yang2025qwen2}, \textbf{GPT-4o} \cite{bubeck2023sparks}, and \textbf{DeepSeek-V3} \cite{liu2024deepseek} are used for zero-shot prediction.
We design a unified prompt template:

\noindent\textit{Given a conversation: <INPUT>. Please analyze the emotion of each utterance in the conversation. The emotions are included in <LABEL SET>.}

\begin{table}[t]\small
\centering
\belowrulesep=0pt
\aboverulesep=0pt
\tabcolsep=2pt
\begin{tabular}{c|cccccc}
\Xhline{0.9pt}
\multirow{2}{*}{\textbf{Models}} & \multicolumn{3}{c}{$\bm{\mathcal{E}}\rightarrow \bm{\mathcal{I}}$ } & \multicolumn{3}{c}{$\bm{\mathcal{M}}\rightarrow \bm{\mathcal{I}}$ } \\
\cmidrule(lr){2-4}\cmidrule(lr){5-7}
                        & \textit{wP.}     & \textit{wR.}    & \textit{wF1.} & \textit{wP.}     & \textit{wR.}    & \textit{wF1.} \\ \hline
                        DialogueGCN& 6.83  & 4.55 & \cellcolor[HTML]{EFEFEF}5.84 & 5.61  & 3.48 & \cellcolor[HTML]{EFEFEF}4.71\\ 
                        DialogueCRN& 7.48  & 6.48 & \cellcolor[HTML]{EFEFEF}6.51  & 7.08  &     5.16  &\cellcolor[HTML]{EFEFEF} 6.44 \\
                        DualGAT& 9.08  & 5.58 &  \cellcolor[HTML]{EFEFEF}7.49  &  7.11      & 5.14      & \cellcolor[HTML]{EFEFEF}6.12 \\
                        CPTC& 17.10     & 10.82  & \cellcolor[HTML]{EFEFEF}14.58  &  13.93      & 10.14      &\cellcolor[HTML]{EFEFEF}11.13 \\
                        SACL-LSTM&	33.05&	24.50&	\cellcolor[HTML]{EFEFEF}20.55&	36.39&19.25&\cellcolor[HTML]{EFEFEF}19.90 \\
                        SCCL&	33.07&	24.56&	\cellcolor[HTML]{EFEFEF}21.21&	36.11&	18.46&	\cellcolor[HTML]{EFEFEF}19.59 \\
                        EACL& 36.10   & 27.42  & \cellcolor[HTML]{EFEFEF}23.89  & 37.00   & 19.53  &\cellcolor[HTML]{EFEFEF}20.79 \\ 
                        \hline
                        Llama-3.1-8b& 38.17	&20.30	&\cellcolor[HTML]{EFEFEF}23.63	&44.33	&30.61	&\cellcolor[HTML]{EFEFEF}24.79  \\
                        Qwen-2.5-7b& 39.34	&21.58	&\cellcolor[HTML]{EFEFEF}24.20	&45.15	&31.37	&\cellcolor[HTML]{EFEFEF}25.66 \\ 
                        GPT-4o& 41.27  & 27.42  & \cellcolor[HTML]{EFEFEF}24.88  & 45.39  & 31.73  & \cellcolor[HTML]{EFEFEF}26.10  \\ 
                        DeepSeek-V3& \underline{42.55}	&\underline{27.69}	&\cellcolor[HTML]{EFEFEF}\underline{25.69}	&\underline{46.38}	&\underline{32.05}	&\cellcolor[HTML]{EFEFEF}\underline{26.26}  \\ \hline 
                        ProEmoTrans (Ours)& \textbf{47.80}  & \textbf{32.95} & \cellcolor[HTML]{EFEFEF}\textbf{37.27} & \textbf{47.11}  & \textbf{30.90}  & \cellcolor[HTML]{EFEFEF}\textbf{32.36} \\ \Xhline{0.9pt}
\end{tabular}
\begin{tabular}{c|cccccc}
\Xhline{0.9pt}
\multirow{2}{*}{\textbf{Models}} & \multicolumn{3}{c}{$\bm{\mathcal{I}}\rightarrow \bm{\mathcal{E}}$ } & \multicolumn{3}{c}{$\bm{\mathcal{M}}\rightarrow \bm{\mathcal{E}}$ }\\
\cmidrule(lr){2-4}\cmidrule(lr){5-7}
                        & \textit{wP.}     & \textit{wR.}    & \textit{wF1.} & \textit{wP.}     & \textit{wR.}    & \textit{wF1.} \\ \hline
                        DialogueGCN& 7.12  & 2.54 & \cellcolor[HTML]{EFEFEF}2.94 & 6.19  & 1.34 & \cellcolor[HTML]{EFEFEF}1.74  \\ 
                        DialogueCRN& 4.32  & 3.27  & \cellcolor[HTML]{EFEFEF}3.29  &  5.52  &  1.23  & \cellcolor[HTML]{EFEFEF}2.09  \\
                        DualGAT& 9.53 & 3.92 &\cellcolor[HTML]{EFEFEF} 4.35  & 3.71  & 1.26  & \cellcolor[HTML]{EFEFEF}1.96 \\
                        CPTC & 7.06  & 4.19  & \cellcolor[HTML]{EFEFEF}5.16 & 3.94   & 1.37 &\cellcolor[HTML]{EFEFEF} 2.41 \\
                        SACL-LSTM &	15.52&	16.49&	\cellcolor[HTML]{EFEFEF}15.34&	14.25&	8.85&	\cellcolor[HTML]{EFEFEF}10.07\\
                        SCCL&14.29	&15.44&	\cellcolor[HTML]{EFEFEF}14.71&	15.00&	9.67&	\cellcolor[HTML]{EFEFEF}11.31\\
                        EACL& 17.48  & 18.01 & \cellcolor[HTML]{EFEFEF}17.36 &  16.36 & 9.74   & \cellcolor[HTML]{EFEFEF}12.39\\ 
                        \hline
                        Llama-3.1-8b &\underline{31.32}	&22.18	&\cellcolor[HTML]{EFEFEF}24.10	&20.11	&\underline{16.83}	&\cellcolor[HTML]{EFEFEF}16.42  \\
                        Qwen-2.5-7b& 31.08	&22.47	&\cellcolor[HTML]{EFEFEF}24.05	&21.17	&16.71	&\cellcolor[HTML]{EFEFEF}17.09\\ 
                        GPT-4o& 31.14   & 21.61  & \cellcolor[HTML]{EFEFEF}\underline{24.51}  & 20.71 & 16.32   & \cellcolor[HTML]{EFEFEF}18.25 \\ 
                        DeepSeek-V3& 31.01	&\underline{23.27}	&\cellcolor[HTML]{EFEFEF}24.10	&\underline{22.91}	&16.27	&\cellcolor[HTML]{EFEFEF}\underline{18.68} \\\hline 
                        ProEmoTrans (Ours)&\textbf{31.36}  & \textbf{27.67}  & \cellcolor[HTML]{EFEFEF}\textbf{28.34} & \textbf{24.98}  & \textbf{19.07} &  \cellcolor[HTML]{EFEFEF}\textbf{20.73}\\ \Xhline{0.9pt}
\end{tabular}
\begin{tabular}{c|cccccc}
\Xhline{0.9pt}
\multirow{2}{*}{\textbf{Models}} & \multicolumn{3}{c}{$\bm{\mathcal{I}}\rightarrow \bm{\mathcal{M}}$ } & \multicolumn{3}{c}{$\bm{\mathcal{E}}\rightarrow \bm{\mathcal{M}}$ }\\
\cmidrule(lr){2-4}\cmidrule(lr){5-7}
                        & \textit{wP.}     & \textit{wR.}    & \textit{wF1.} & \textit{wP.}     & \textit{wR.}    & \textit{wF1.} \\ \hline
                        DialogueGCN& 5.76  & 3.12  & \cellcolor[HTML]{EFEFEF}4.44  & 5.42  & 1.99  & \cellcolor[HTML]{EFEFEF}2.67 \\ 
                        DialogueCRN& 7.46  & 4.00   & \cellcolor[HTML]{EFEFEF}5.13  & 6.93   & 3.15  & \cellcolor[HTML]{EFEFEF}3.95 \\
                        DualGAT & 8.20  & 4.12   & \cellcolor[HTML]{EFEFEF}5.07   & 6.23   & 2.08      &  \cellcolor[HTML]{EFEFEF}2.94 \\
                        CPTC & 19.51   & 5.65  &\cellcolor[HTML]{EFEFEF} 8.13 & 13.69  & 4.30  & \cellcolor[HTML]{EFEFEF}6.40  \\
                        SACL-LSTM &	31.60&	19.29&	\cellcolor[HTML]{EFEFEF}25.60	&29.55	&22.28&	\cellcolor[HTML]{EFEFEF}25.48\\
                        SCCL&	31.05&	19.39&	\cellcolor[HTML]{EFEFEF}25.14&	28.81&	21.02&	\cellcolor[HTML]{EFEFEF}24.32\\
                        EACL& \underline{33.32} & 20.27  & \cellcolor[HTML]{EFEFEF}26.29  & 31.58  & 23.52 &  \cellcolor[HTML]{EFEFEF}26.95\\ 
                        \hline
                        Llama-3.1-8b &31.08	&22.47	&\cellcolor[HTML]{EFEFEF}24.05	&32.81	&25.36	&\cellcolor[HTML]{EFEFEF}27.73  \\
                        Qwen-2.5-7b & 30.50	&43.80	&\cellcolor[HTML]{EFEFEF}35.12	&34.72	&26.96	&\cellcolor[HTML]{EFEFEF}29.76\\ 
                        GPT-4o  & 29.90   &  \underline{45.65}& \cellcolor[HTML]{EFEFEF}\underline{35.28}  & \underline{34.85}  & 25.74   & \cellcolor[HTML]{EFEFEF}29.35 \\ 
                        DeepSeek-V3& \underline{31.85}	&44.67	&\cellcolor[HTML]{EFEFEF}35.15	&34.76	&\underline{27.19}	&\cellcolor[HTML]{EFEFEF}\underline{29.76} \\ \hline 
                        ProEmoTrans (Ours) & \textbf{35.74}  & \textbf{45.32}  & \cellcolor[HTML]{EFEFEF}\textbf{38.59} & \textbf{36.30}  & \textbf{36.02}   & \cellcolor[HTML]{EFEFEF}\textbf{35.64}  \\ \Xhline{0.9pt}
\end{tabular}
\caption{The overall performance of all the compared baselines and our ProEmoTrans on benchmark datasets. Here \textit{wP.}, \textit{wR.}, and \textit{wF1.} denote weighted-averaged precision, recall, and F1 score.}
\label{tab:main}
\end{table}

\section{Results and Analysis}

\begin{table*}[t]\small
\centering
\begin{tabular}{l|cccccc|c}
\Xhline{0.9pt}
\multicolumn{1}{c|}{\textbf{Models}}       & $\bm{\mathcal{E}}\rightarrow\bm{\mathcal{I}}$    &  $\bm{\mathcal{M}}\rightarrow\bm{\mathcal{I}}$      &  $\bm{\mathcal{I}}\rightarrow\bm{\mathcal{E}}$     &  $\bm{\mathcal{M}}\rightarrow\bm{\mathcal{E}}$      &  $\bm{\mathcal{I}}\rightarrow\bm{\mathcal{M}}$     &  $\bm{\mathcal{E}}\rightarrow\bm{\mathcal{M}}$  & Average     \\ \hline
Proposed ProEmoTrans   & \textbf{37.27} & \textbf{32.36}  & \textbf{28.34} & \textbf{20.73} & \textbf{38.59} & \textbf{35.64} & \textbf{32.16} \\ \hline
\quad - w/o \textit{LED} & 27.68  &  7.28   &  24.06 & 6.31  &  22.59 & 19.22 & 17.86 ($14.30^\downarrow$)    \\ 
\quad - w 1 \textit{Desc.}  & 30.46 &   9.90  & 25.06    &  8.74  & 24.68  &  25.26 & 20.68 ($11.48^\downarrow$) \\
\quad - w 3 \textit{Desc.}  & 37.56  & 33.03  & 28.39   &  21.22  &  37.89  &  36.82  & 32.49 ($\textbf{0.33}^\uparrow$)  \\  \hline
\quad - w/o \textit{GSA}  & 36.89 & 31.40   & 27.00  & 19.37& 37.68 &     34.26  & 31.10 ($1.06^\downarrow$) \\ 
\quad - w \textit{SA}   & 36.27   & 30.78  & 26.47   & 19.82 &37.01  &  33.45 &30.63 ($1.53^\downarrow$) \\
\hline
\quad - w/o \textit{CRF}    & 31.22  & 19.20 & 18.29  & 16.59   &  33.82 & 32.27 &  25.23 ($6.93^\downarrow$) \\ \Xhline{0.9pt}
\end{tabular}
\caption{Ablation and comparison results for key components. Here 1 \textit{Desp.} and 3 \textit{Desp.} denote the number of generated descriptions in LED. \textit{SA} denotes replacing GSA with the original self-attention mechanism.
}
\label{tab:ablation}
\end{table*}

\subsection{Main Results}

The overall performance on the three datasets is reported in Table \ref{tab:main}.
We have the following observations:
Our ProEmoTrans outperforms all other models by a significant margin. Compared to the best baseline DeepSeek-V3, ProEmoTrans achieved improvements in the weighted-averaged F1 score of 11.58\%, 6.1\%, 4.24\%, 2.05\%, 3.44\%, and 5.88\% across six different dataset settings. This demonstrates that our ProEmoTrans exhibits strong performance.
The feature-based methods DialogueGCN, DialogueCRN, and DualGAT perform poorly due to their excessive parameter modules, which make them prone to overfitting on seen emotions.
Few-shot model CPTC also shows inefficient recognition of unseen emotions.
The contrastive-based methods SACL-LSTM, SCCL, and EACL focus on improving the distinguishability of different emotions. Learning differentiated emotional prototypes helps them perform better on the UERC task than other supervised methods.
LLMs outperform other baselines with their rich prior knowledge.
To investigate how our model improves performance compared to GPT-4o,  we provide a more in-depth discussion in the fine-grained analysis (Section \ref{sec:fine-grained}).

\subsection{Ablation Study}
We conduct ablation studies to investigate the effectiveness of the key components in our method. The results are shown in Table \ref{tab:ablation}.

\textbf{-w/o \textit{LED}} denotes removing the LED module and directly using dictionary definitions as its description.
It is evident that removing the LED results in a significant 14.3\% drop in the model's average wF1 score, highlighting the importance of descriptive information in enhancing emotion representation.
In the original model, we use two descriptions (\textit{2 Desc.}) to help the model fully capture the emotional semantics. 
To investigate the impact of the number of generated descriptions, we conduct experiments comparing the model's performance with different numbers of descriptions. As shown in Table \ref{tab:ablation}, with one description (-w 1 \textit{Desc.}), the average wF1 increases by 2.82\% compared to \textit{no Desp}. However, it still shows an 11.48\% drop compared to the original 2 \textit{Desc.}.
With three descriptions (-w 3 \textit{Desc.}), the average wF1 only slightly increases by 0.33\%. 
This indicates that 2 \textit{Desc.} are sufficient for the model to fully capture the semantic meaning.

\textbf{-w/o \textit{GSA}} denotes removing the GSA mechanism and directly using $\bm{H}$ from Eq. (\ref{eq3})  as the final utterance representations. 
This led to a decrease of 1.06\% in the average wF1, demonstrating the positive role of the GSA mechanism in enhancing utterance representations.
Since the GSA mechanism benefits from aggregating highly relevant information while reducing the negative impact of distant utterances, we further compare it with using the self-attention mechanism (\textit{SA}) alone. 
The results show that the performance drops by 1.53\%, and it even performs 0.47\% worse than when no mechanism was used (-w/o \textit{GSA}). 
This demonstrates that directly using \textit{SA} for utterance representation learning has a detrimental effect, with the negative impact stemming from distant noise.
%

\textbf{-w/o \textit{CRF}} denotes removing the CRF layer and the AVD algorithm, and during the inference phase, it directly uses $h^{see}_i$ and $h^{uns}_j$ for nearest neighbor search as specified in Eq. (\ref{eq15}).
The results show a decrease of 6.93\% in average wF1, which demonstrates that the AVD algorithm, by leveraging the emotion transition dependencies learned by the CRF layer, plays a crucial role in enhancing the model's performance.

\subsection{Hyperparameter Sensitivity}
\begin{figure}[t]
    \centering
    \includegraphics[width=0.48\textwidth]{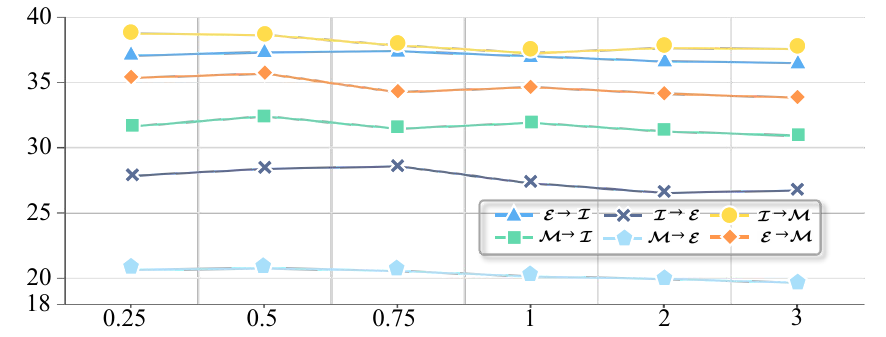}
    \vspace{-0.7cm}
    \caption{Effects of $\sigma$.}
    \label{fig:effects}
\end{figure}
The variance $\sigma$ in the GSA mechanism controls the attention range. 
To study the impact of $\sigma$ on performance, we conducted a sensitivity analysis, as shown in Figure \ref{fig:effects}.
It can be observed that the best performance is achieved when $\sigma$ is set to 0.5. As $\sigma$ increases, the performance gradually decreases and converges. In fact, as $\sigma$ grows, the Gaussian Self-Attention mechanism gradually degenerates into a standard self-attention mechanism.

\begin{table}[t]\small
\centering
\begin{tabular}{ccc}
\Xhline{0.9pt}
Model             & Performance & Inference Costs \\\hline
bert-base-uncased &  32.16    &   6.21 /ms    \\
roberta-base    &  33.02   &   6.34 /ms    \\ 
bert-large-uncased &  34.48    &  9.12 /ms    \\
roberta-large    & 34.83  &   9.26 /ms   \\ \Xhline{0.9pt}
\end{tabular}
\caption{Performance (wF1.) and computation cost (/ms) with different language models}
\label{tab:performance}
\end{table}

\begin{figure*}[ht]
    \centering
    \begin{minipage}{0.32\textwidth}
        \centering
        \includegraphics[width=\textwidth]{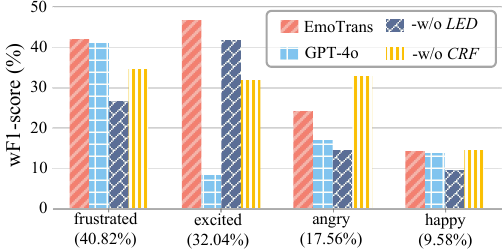}
        \subcaption{$\bm{\mathcal{E}}\rightarrow \bm{\mathcal{I}}$}
    \end{minipage}%
    \begin{minipage}{0.32\textwidth}
        \centering
        \includegraphics[width=\textwidth]{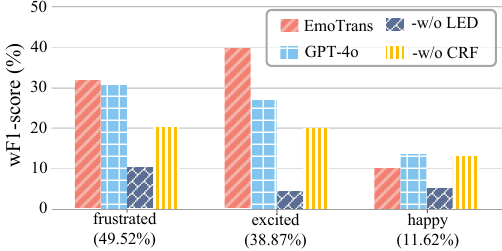}
        \subcaption{$\bm{\mathcal{M}}\rightarrow \bm{\mathcal{I}}$}
    \end{minipage}%
    \begin{minipage}{0.32\textwidth}
        \centering
        \includegraphics[width=\textwidth]{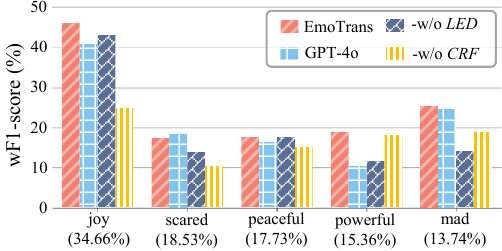}
        \subcaption{$\bm{\mathcal{I}}\rightarrow \bm{\mathcal{E}}$}
    \end{minipage}

    \vspace{1ex}

    \begin{minipage}{0.32\textwidth}
        \centering
        \includegraphics[width=\textwidth]{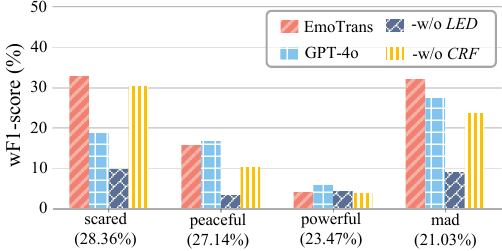}
        \subcaption{$\bm{\mathcal{M}}\rightarrow \bm{\mathcal{E}}$}
    \end{minipage}%
    \begin{minipage}{0.32\textwidth}
        \centering
        \includegraphics[width=\textwidth]{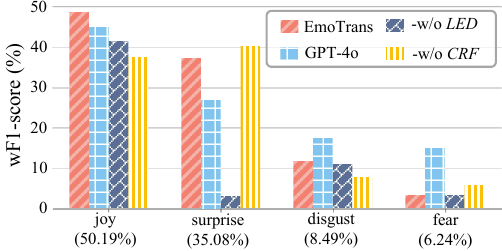}
        \subcaption{$\bm{\mathcal{I}}\rightarrow \bm{\mathcal{M}}$}
    \end{minipage}%
    \begin{minipage}{0.32\textwidth}
        \centering
        \includegraphics[width=\textwidth]{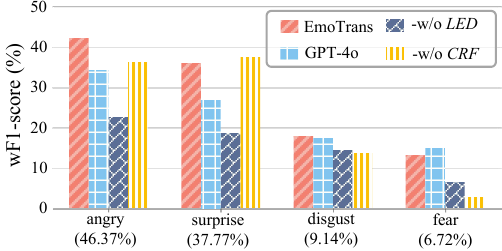}
        \subcaption{$\bm{\mathcal{E}}\rightarrow \bm{\mathcal{M}}$}
    \end{minipage}
    \vspace{-0.2cm}
    \caption{Fine-grained analysis of different methods, with the proportion of unseen emotions also presented.}
    \label{fig:fine-grained}
\end{figure*}

\subsection{Average Performance and Computation Cost with Different Language Models}
To investigate the effect of using different pre-trained language models and the corresponding computation costs, we conduct experiments and record the average performance and inference costs in Table \ref{tab:performance}.
Using roberta-base \cite{DBLP:journals/corr/abs-1907-11692} improves the model's average performance by 0.86\%. 
With the larger versions, Bert and Roberta improve the model's average performance by 2.32\% and 1.81\%, respectively. 
However, the average inference time per sample increases by 2.91 ms and 2.92 ms, respectively.

\subsection{Fine-grained Analysis}
\label{sec:fine-grained}
As shown in Figure \ref{fig:fine-grained}, we conduct an experiment to demonstrate the fine-grained performance of different methods.
Comparing the performance of ProEmoTrans and GPT-4o, we can observe that ProEmoTrans performs better in most unseen emotions. 
However, as the emotion proportion decreases, ProEmoTrans shows a more noticeable decline in performance.
We believe this is due to GPT-4o relying on prior knowledge, while ProEmoTrans depends on the quality of prototype-based representation learning, which makes it more sensitive to the distribution of categories.

Removing the LED (-w/o \textit{LED}) causes a performance drop across all unseen emotions, to varying degrees, highlighting the LED's comprehensive contribution.
Similarly, removing the CRF (-w/o \textit{CRF}) also leads to a nearly overall performance decline, but in some cases, it improves performance.
For example, in subplot (f), it leads to a 2.45\% increase for \textit{surprise}.
This suggests that while the CRF layer optimizes global performance, it may not be ideal for certain local categories.

\begin{figure}[t]
    \centering
    \begin{minipage}{0.25\textwidth}
        \centering
        \includegraphics[width=\textwidth]{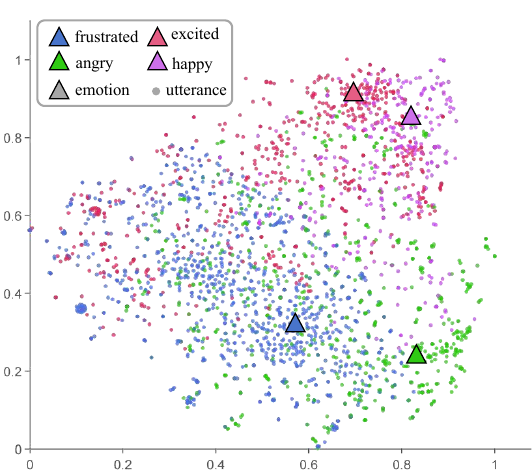}
        \vspace{-0.5cm}
        \subcaption{Proposed ProEmoTrans}
    \end{minipage}%
    \begin{minipage}{0.25\textwidth}
        \centering
        \includegraphics[width=\textwidth]{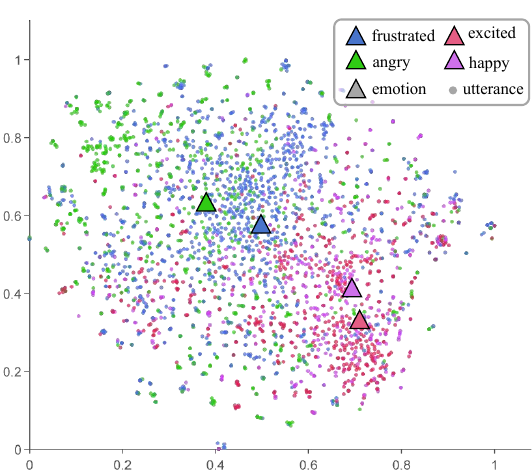}
        \vspace{-0.5cm}
        \subcaption{-w/o \textit{CRF}}
    \end{minipage}
    \vspace{-0.2cm}
    \caption{t-SNE visualization of utterance and emotion embeddings in $\bm{\mathcal{E}}\rightarrow \bm{\mathcal{I}}$ datasets.}
    \label{fig:visua}
\end{figure}

\subsection{Visualization}

To provide more interpretability, we visualize the embedding space of utterances and unseen emotions on $\bm{\mathcal{E}}\rightarrow \bm{\mathcal{I}}$ datasets using t-SNE \cite{van2008visualizing}, as shown in Figure \ref{fig:visua}.
First, we find that positive emotions (\textit{excited} and \textit{happy}) are farther apart from negative emotions (\textit{frustrated} and \textit{angry}), while emotions of the same polarity are closer to each other, which aligns with our intuition.
Next, comparing subfigures (a) and (b), we can see that adding the CRF layer enhances the distinguishability of utterance and emotion embeddings, demonstrating the positive impact of the CRF layer and AVD algorithms in our method.

\begin{figure}[t]
    \centering
    \includegraphics[width=0.3\textwidth]{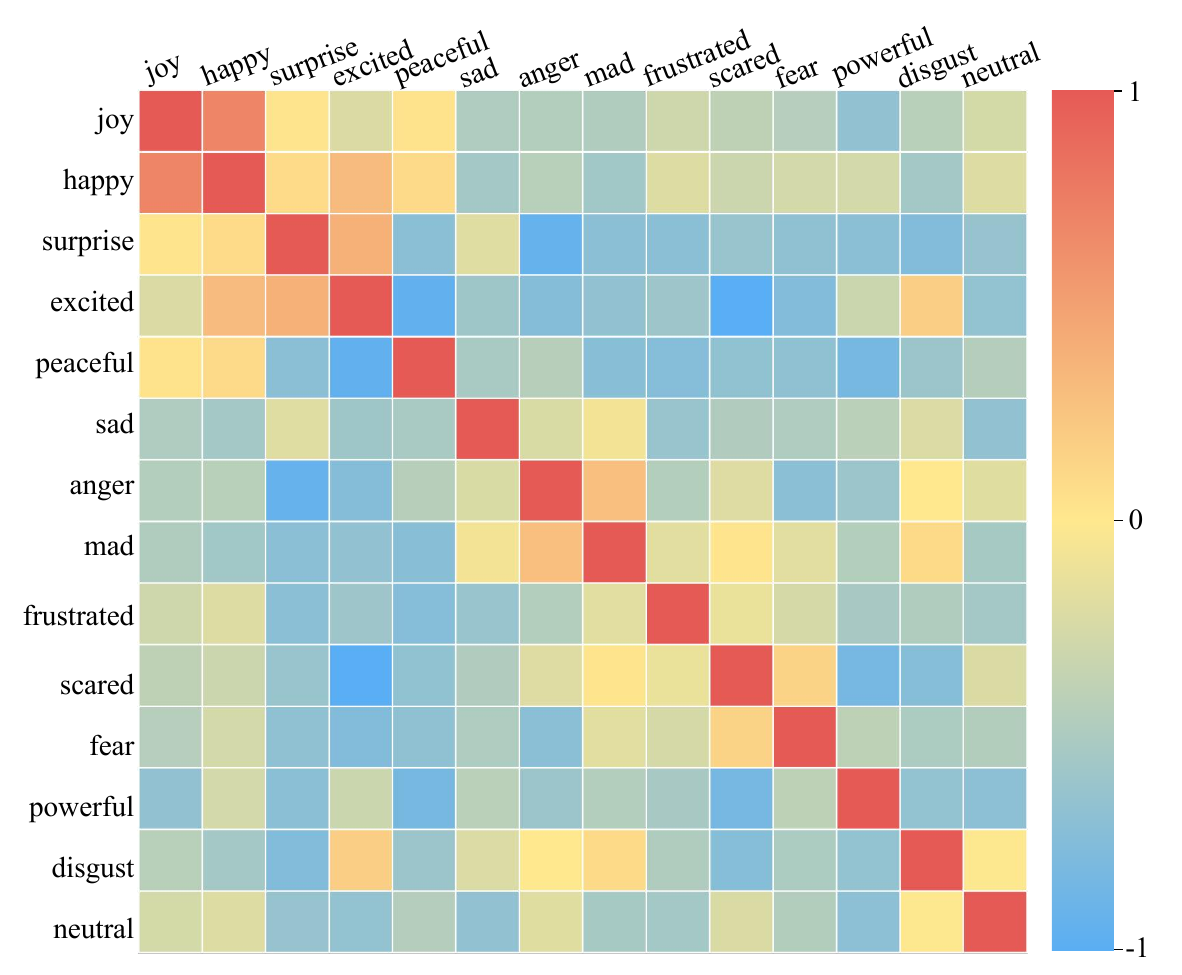}
    \vspace{-0.2cm}
    \caption{Heatmap of emotion prototype similarities.}
    \label{fig:heatmap}
\end{figure}

We also collect all the emotion prototype embeddings and compute their cosine similarities. The resulting heatmap is shown in Figure \ref{fig:heatmap}.
It can be observed that, first, the cosine similarity is higher between similar emotions (e.g., \textit{happy} and \textit{joy}). Second, there is a more pronounced difference in similarity between positive and negative emotions.

\begin{table*}[t] \small
\centering
\begin{tabular}{l|cc|cc|cc|c}
\Xhline{0.9pt}
\multicolumn{1}{c|}{\textbf{Models}}       & $\bm{\mathcal{E}}\rightarrow\bm{\mathcal{I}}$    &  $\bm{\mathcal{M}}\rightarrow\bm{\mathcal{I}}$      &  $\bm{\mathcal{I}}\rightarrow\bm{\mathcal{E}}$     &  $\bm{\mathcal{M}}\rightarrow\bm{\mathcal{E}}$      &  $\bm{\mathcal{I}}\rightarrow\bm{\mathcal{M}}$     &  $\bm{\mathcal{E}}\rightarrow\bm{\mathcal{M}}$  & Average     \\ \hline
Contrastive Similarity (Ours)   & \textbf{37.27} & \textbf{32.36}  & \textbf{28.34} & \textbf{20.73} & \textbf{38.59} & \textbf{35.64} & \textbf{32.16} \\ \hline
\quad - w \textit{Euclidean distances} & 35.12	&30.03	&27.23	&18.89	&37.77	&34.07	&30.52    \\ 
\quad - w \textit{Cosine similarity}   & 36.45	&31.91	&27.67	&20.10	&37.85	&34.98	&31.49 \\
\quad - w \textit{Dot Product}    & 35.78	&30.56	&27.34	&19.52	&36.29	&34.27	&30.63 \\ \Xhline{0.9pt}
\end{tabular}
\caption{The results of comparing contrastive similarity with other similarity metrics.}
\label{tab:Contrastive}
\end{table*}

\subsection{More Additional Experiments}

\subsection{Analysis on Contrastive Similarity}
The contrastive similarity \cite{oord2018representation} can effectively measure the difference between two embeddings. To validate its effectiveness, we conducted experiments comparing it with other similarity metrics. The results are shown in Table~\ref{tab:Contrastive}.
When using Euclidean distance, cosine similarity, and dot product, the model's performance decreased by 1.64\%, 0.67\%, and 1.53\%, respectively, which proves the effectiveness of contrastive similarity.

\subsubsection{Few-shot Performance}
Our model can also be used for few-shot prediction without any modifications. 
To investigate the performance of our model in the few-shot setting, we conducted experiments as shown in Table \ref{tab:fewshot}.
To ensure a fair comparison with the baselines, we follow the 16-shot setting and use weighted macro-F1 as the evaluation metric.
The applied baselines include: 
\textbf{KEY} \cite{zhong-etal-2019-knowledge} addresses ERC tasks by utilizing external knowledge bases.
%
\textbf{TUCORE-GCN} \cite{lee-choi-2021-graph} and \textbf{EmotionFlow} \cite{DBLP:conf/icassp/SongZZHH22} are GCN-based and RNN-based ERC model, respectivly.
\textbf{SPCL} \cite{song-etal-2022-supervised} uses supervised contrastive learning to address the class imbalance problem in ERC.
\textbf{CTPT} \cite{xu-etal-2023-efficient} is introduced in Section \ref{sec:baseline}.
According to the results, our ProEmoTrans outperforms the best baseline, CTPT, by 1.89\%, 1.38\%, and 2.01\% on the three datasets, respectively. This demonstrates that our model also performs excellently in the few-shot setting.

\begin{table}[t] \small
\centering
\tabcolsep=10pt
\begin{tabular}{c|ccc}
\Xhline{0.9pt}
\multicolumn{1}{c|}{\textbf{Models}}             & $\bm{\mathcal{E}}$ & $\bm{\mathcal{I}}$ & $\bm{\mathcal{M}}$ \\\hline
KET & 13.12  &   16.46 & 8.97   \\
TUCORE-GCN    &  13.11   & 15.27& 25.96   \\ 
EmotionFlow &  14.65    &  16.99 & 29.34   \\
SPCL    & 14.99  &  18.73& 29.41\\ 
CTPT  & \underline{20.57}  & \underline{31.82} & \underline{31.28}\\ \hline
ProEmoTrans (Ours) & \textbf{22.46}  & \textbf{33.20} & \textbf{33.29}\\ \hline
\Xhline{0.9pt}
\end{tabular}
\caption{Performance of different ERC datasets under the few-shot settings (16-shot). All the baseline results are retrieved from \citet{xu-etal-2023-efficient}. We \textbf{bolded} the best result and \underline{underline} the second best.}
\label{tab:fewshot}
\end{table}

\section{Conclusion}
In this paper, we propose a simple and effective method named ProEmoTrans for the newly proposed UERC task.
First, we introduce an LLM-enhanced Emotion Description module to enhance emotion prototype learning.
Next, a parameter-free Gaussian Self-Attention mechanism is designed to aggregate useful information from the conversation while filtering out noise. 
This mechanism can learn inter-utterance relations and prevent overfitting that could arise from parameter training.
Finally, we propose an Attention Viterbi Decoding algorithm to transfer the useful seen emotion dependencies learned during training to unseen emotions. 
Extensive experiments on three datasets validate the effectiveness of our approach and the individual modules we designed.
In future work, our goal is to further optimize prototype representations.

\section{Limitations}
Our LLM prompt templates rely on manual design, and their effectiveness has not been verified with more complex emotions. Developing automated prompt-tuning templates would be an interesting avenue for exploration. 
Additionally, our approach focuses solely on the text modality and does not incorporate multi-modal information, such as facial expressions, which could provide valuable additional information.

\section{Acknowledgments}
This research is supported by the National Key R\&D Program of China (No. 2023YFC3303800), NSFC through grants 62322202, 62441612 and 62476163, Beijing Natural Science Foundation through grant L253021, Local Science and Technology Development Fund of Hebei Province Guided by the Central Government of China through grants 246Z0102G and 254Z9902G, the “Pionee” and “Leading Goose” R\&D Program of Zhejiang through grant 2025C02044, Hebei Natural Science Foundation through grant F2024210008, and the Guangdong Basic and Applied Basic Research Foundation through grant 2023B1515120020.

\bibliography{custom}

\appendix
\newpage

\section{More Details of Experiments Settings}
\label{sec:appendixa2}

\subsection{Datasets}
\label{sec:app-dataset}
\begin{table}[h]\small
\tabcolsep=3pt
\centering
\begin{tabular}{c|ccc}
\Xhline{0.9pt}
                       \multirow{2}{*}{Source} & \multicolumn{3}{c}{Target} \\ \cline{2-4} 
                        & $\bm{\mathcal{I}}$& $\bm{\mathcal{E}}$ & $\bm{\mathcal{M}}$      \\ \hline 
\multicolumn{1}{c|}{$\bm{\mathcal{I}}$} & / & \textit{po}, \textit{pe}, \textit{sc}, \textit{jo}, \textit{ma}  & \textit{su}, \textit{di}, \textit{fe}, \textit{jo}  \\

\multicolumn{1}{c|}{$\bm{\mathcal{E}}$} & \textit{ex}, \textit{fr}, \textit{ha}, \textit{an}  & /  &\textit{su}, \textit{di}, \textit{fe}, \textit{an} \\
\multicolumn{1}{c|}{$\bm{\mathcal{M}}$} &  \textit{ex}, \textit{fr}, \textit{ha} &   \textit{po}, \textit{pe}, \textit{sc}, \textit{ma} & /  \\
\Xhline{0.9pt}
\end{tabular}
\caption{Statistics of unseen emotions under different source and target settings. We use the first two letters to denote the emotions in Table \ref{tab:statistics}, for example, \textit{po} stands for \textit{powerful}.}
\label{tab:statistics2}
\end{table}
The statistics of the unseen emotions under different source and target settings are shown in Table \ref{tab:statistics2}.
For example, if we chose $\bm{\mathcal{I}}$ as source dataset and $\bm{\mathcal{E}}$ as target dataset, the unseen emotions are \textit{powerful}, \textit{peaceful}, \textit{scared}, \textit{joy}, and \textit{mad}.

\subsection{Baselines}
\label{sec:app-Baselines}
The details of baselines are as follows:
\begin{itemize}[leftmargin=1em]
\item \textbf{DialogueGCN} \cite{ghosal-etal-2019-dialoguegcn} uses a GCN to model the inter-utterance dependency.
\item \textbf{DialogueCRN} \cite{hu-etal-2021-dialoguecrn} is one of the best RNN-based ERC models. They design multiple rounds of reasoning modules to extract and integrate emotional cues.
\item \textbf{DualGAT} \cite{zhang-etal-2023-dualgats} introduces a Dual Graph Attention Network to capture complex dependencies of discourse structure and speaker-aware context.
\item \textbf{SACL-LSTM} \cite{hu-etal-2023-supervised} proposes a supervised adversarial contrastive learning method for learning class-spread structured representations.
\item \textbf{SCCL} \cite{10040720} proposes a supervised cluster-level contrastive learning method to incorporate measurable emotion prototypes.
\item \textbf{EACL} \cite{yu-etal-2024-emotion} proposes an emotion-anchored contrastive learning framework, which generates more distinguishable utterance representations for similar emotions.
\item \textbf{CPTC} \cite{xu-etal-2023-efficient} leverages sharable cross-task knowledge from the source task to improve few-shot performance.
\end{itemize}
We made the necessary modifications for each baseline to enable zero-shot prediction.

\section{More Additional Experiments}
\label{sec:appendix2}

\begin{table*}[th]\small
\centering
\tabcolsep=2.3pt
\begin{tabular}{l|cc|cc|cc|c}
\Xhline{0.9pt}
\multicolumn{1}{c|}{\textbf{Models}}       & $\bm{\mathcal{E}}\rightarrow\bm{\mathcal{I}}$    &  $\bm{\mathcal{M}}\rightarrow\bm{\mathcal{I}}$      &  $\bm{\mathcal{I}}\rightarrow\bm{\mathcal{E}}$     &  $\bm{\mathcal{M}}\rightarrow\bm{\mathcal{E}}$      &  $\bm{\mathcal{I}}\rightarrow\bm{\mathcal{M}}$     &  $\bm{\mathcal{E}}\rightarrow\bm{\mathcal{M}}$  & Average     \\ \hline
Proposed ProEmoTrans   & \textbf{37.27} & \textbf{32.36}  & \textbf{28.34} & \textbf{20.73} & \textbf{38.59} & \textbf{35.64} & \textbf{32.16} \\ \hline
\quad -w \textit{LSTM} & 	9.37 & 	8.82 & 	7.74 & 	6.04 & 	7.70 & 	6.78 & 	7.74 \\ 
\quad -w \textit{GCN} & 	8.42 & 	8.10 & 	6.47 & 	5.19 & 7.63 & 	5.97 & 	6.96 \\ 
\quad -w \textit{GAT} & 	7.57 & 	8.93 & 	6.19 & 	5.65 & 	7.11 & 	6.08 & 	6.92 \\ 
\Xhline{0.9pt}
\end{tabular}
\caption{Comparative experiments by replacing the GSA module with LSTM, GCN, and GAT.}
\label{tab:ablationxx}
\end{table*}

\subsection{Results with Parameterized Modules}
\label{sec:appendixa1}
In supervised settings, previous methods have designed various parameterized modules to help learn better utterance representations. 
In the zero-shot setting, to validate their effectiveness, we conduct comparative experiments by replacing the Gaussian Self-Attention module in our model with LSTM, GCN, and GAT. The experimental results are shown in Table \ref{tab:ablationxx}.
It can be observed that the performance is quite weak, which proves that overfitting due to the parameter module severely hinders the generalization performance.

\begin{table*}[t]\small
\centering
\tabcolsep=2.8pt
\begin{tabular}{l|cccccc|c}
\Xhline{0.9pt}
\multicolumn{1}{c|}{\textbf{Models}}       & $\bm{\mathcal{E}}\rightarrow\bm{\mathcal{I}}$    &  $\bm{\mathcal{M}}\rightarrow\bm{\mathcal{I}}$      &  $\bm{\mathcal{I}}\rightarrow\bm{\mathcal{E}}$     &  $\bm{\mathcal{M}}\rightarrow\bm{\mathcal{E}}$      &  $\bm{\mathcal{I}}\rightarrow\bm{\mathcal{M}}$     &  $\bm{\mathcal{E}}\rightarrow\bm{\mathcal{M}}$  & Average     \\ \hline
Proposed ProEmoTrans   & \textbf{37.27} & \textbf{32.36}  & \textbf{28.34} & \textbf{20.73} & \textbf{38.59} & \textbf{35.64} & \textbf{32.16} \\ \hline
DeepSeek-V3 &25.69	&26.26	&24.10	&18.68	&35.15	&29.76 & 26.61 ($5.55^\downarrow$) \\
\quad -w utterance-level &30.06	&31.52	&23.27	&19.04	&34.86	&29.45 & 28.03 ($4.13^\downarrow$)\\\hline
GPT-4o &24.88	&26.10	&24.51	&18.25	&35.28	&29.35 &26.40 ($5.76^\downarrow$)\\
\quad -w utterance-level &30.24	&31.25	&24.22	&18.48	&35.07	&29.51 &28.13 ($4.03^\downarrow$)\\
\Xhline{0.9pt}
\end{tabular}
\caption{Performance comparison on Zero-Shot ERC at the utterance level.
}
\label{tab:ablation2}
\end{table*}

\subsection{Utterance-level Performance}
We conducted a comparative experiment on zero-shot ERC at the utterance level, with results shown in Table~\ref{tab:ablation2}, where \textbf{-w utterance-level} refers to applying LLM baselines to prompt each individual utterance.
Our experiments uncovered some intriguing findings:
On the longer dialogue dataset (IEMOCAP, avg. length ~52), utterance-level classification significantly outperformed the original conversation-level approach. We believe that excessively long conversations hinder LLM’s emotional analysis capability by overwhelming context processing.
On the other two datasets (avg. lengths 12 and 9), utterance-level performance was slightly lower than conversation-level. We attribute this to the loss of contextual information, which poses challenges for utterances with ambiguous emotional cues or those that are very brief. For example, the utterance "That only took me an hour." was misclassified as joy at the utterance level, but correctly classified as sad at the conversation level when the broader topic (divorce) was considered. Crucially, our method consistently maintains an advantage across different conversation lengths, despite the observed variations in zero-shot LLM classification performance.

\section{Details of LED Generated Descriptions}
\label{sec:appendix}
To eliminate biases introduced by the quality of generated descriptions, we regenerate new descriptions in each of the five random runs. The emotion descriptions generated using the LED module in one of the five runs are shown in Table \ref{tab:desp}.


\begin{table*}[b] \small
\begin{tabularx}{\textwidth}{c|X|X}
\Xhline{0.9pt}
\textbf{Emotion} & \textbf{Description}      & \textbf{LED Generated Description}   \\ \hline
joy     & a feeling of extreme happiness or cheerfulness, especially related to the acquisition or expectation of something good. & 1. When she received the good news, a bright smile spread across her face. 2. At the celebration, laughter filled the room, creating a vibrant atmosphere. \\ \hline
happy & the emotion of being happy; elation.  & 1. She smiled as she read the message, feeling content with the little surprise. 2. As the music played, a warm sense of fulfillment washed over her, making everything else fade away.  \\ \hline
surprise& the feeling that something unexpected has happened. & 1. The unexpected turn of events left him speechless, eyes wide in disbelief. 2. She could hardly believe her ears when she heard the astonishing news.\\ \hline
excited  & having great enthusiasm, passion, and energy. & 1. Her heart raced as she opened the envelope containing the results. 2. He couldn't sit still, eagerly anticipating the start of the event.\\\hline
peaceful & motionless and calm. & 1. The gentle sound of the waves lapping against the shore filled her with calm. 2. Sitting under the shade of the old tree, he felt completely at ease.\\ \hline
sad  & emotionally negative and feeling sorrow. & 1. He stared out the window, his heart heavy with a lingering sense of loss. 2. As she walked through the empty hall, a wave of nostalgia washed over her.\\ \hline
anger  & displaying or feeling anger. & 1. She clenched her fists and glared at him when she heard the unfair criticism. 2. His face turned red as he confronted the person who had betrayed his trust.\\ \hline
mad & feeling annoyed.  & 1. He slammed the door shut, frustration boiling beneath the surface. 2. Her eyes flashed with irritation as she listened to the unfair remarks.\\ \hline
frustrated  & the feeling of annoyance at impossibility from resistance or inability to achieve something. & 1. She let out a sigh and ran her fingers through her hair, feeling exasperated with the situation. 2. He stared at the puzzle pieces scattered on the table, unable to find a solution.\\ \hline
scared & feeling afraid and frightened.  & 1. A cold sweat broke out on his forehead as he heard footsteps behind him in the dark. 2. She held her breath, feeling a knot tighten in her stomach during the thunderstorm.\\ \hline
fear& a strong, unpleasant emotion or feeling caused by actual or perceived danger or threat. & 1. In the dark alley, a sudden noise made his heart race with unease. 2. She felt a chill run down her spine as shadows flickered around her.\\ \hline
powerful&having, or capable of exerting, power or influence. & 1. Standing at the edge of the cliff, she felt an overwhelming sense of strength and determination. 2. The speaker's voice resonated through the hall, commanding everyone’s attention.\\ \hline
disgust& to cause an intense dislike for something. & 1. I couldn't believe it when my teammate ignored my advice during the game. 2. It drove me crazy when the internet kept disconnecting while I was working.\\ \hline
neutral  & neither positive nor negative. & 1. He sat quietly, showing no particular reaction to the events around him. 2. The room was filled with a quiet stillness as everyone focused on their tasks.\\ \hline
\Xhline{0.9pt}

\end{tabularx}
\caption{Details of LED generated descriptions}
\label{tab:desp}
\end{table*}

\end{document}